\documentclass[runningheads,a4paper]{llncs}
\usepackage[utf8]{inputenc}

\usepackage{amssymb}
\usepackage{graphicx}
\usepackage{amsmath}
\usepackage{latexsym}

\usepackage{url}

\usepackage[numbers]{natbib}

\newcommand{\Hh}{{\widehat{H}}}

\newcommand{\argmax}{\mathop{\mathrm{argmax\,}}}
\newcommand{\boldC}{{\boldsymbol{C}}}
\newcommand{\boldp}{{\boldsymbol{p}}}
\newcommand{\boldHh}{{\widehat{\boldH}}}
\newcommand{\boldH}{{\boldsymbol{H}}}
\newcommand{\boldI}{{\boldsymbol{I}}}
\newcommand{\boldK}{{\boldsymbol{K}}}
\newcommand{\boldM}{{\boldsymbol{M}}}
\newcommand{\boldU}{{\boldsymbol{U}}}
\newcommand{\boldalpha}{{\boldsymbol{\alpha}}}
\newcommand{\boldhh}{{\widehat{\boldh}}}
\newcommand{\boldh}{{\boldsymbol{h}}}

\newcommand{\boldomegah}{{\widehat{\boldomega}}}
\newcommand{\boldomega}{{\boldsymbol{\omega}}}
\newcommand{\boldone}{{\boldsymbol{1}}}
\newcommand{\boldphi}{{\boldsymbol{\phi}}}

\newcommand{\boldx}{{\boldsymbol{x}}}
\newcommand{\boldzero}{{\boldsymbol{0}}}
\newcommand{\calN}{{\mathcal{N}}}
\newcommand{\calZ}{{\mathcal{Z}}}
\newcommand{\densitymodel}{\densitysymbol}
\newcommand{\densitysymbol}{p}
\newcommand{\density}{\densitysymbol^\ast}
\newcommand{\hh}{{\widehat{h}}}
\newcommand{\inputdim}{d}

\newcommand{\kernelparameter}{t}
\newcommand{\mathbbR}{\mathbb{R}}
\newcommand{\nclass}{c}
\newcommand{\nsample}{n}
\newcommand{\omegah}{{\widehat{\omega}}}
\newcommand{\ratioh}{\widehat{\ratiosymbol}}
\newcommand{\ratiomodel}{\ratiosymbol}
\newcommand{\ratiosymbol}{r}
\newcommand{\ratio}{\ratiosymbol^\ast}

\begin{document}

\mainmatter

\title{
Semi-Supervised\\
Information-Maximization Clustering
}
\titlerunning{Semi-Supervised Information-Maximization Clustering}

\author{Daniele Calandriello \inst{1} \and Gang Niu \inst{2} \and Masashi Sugiyama \inst{2}}
\institute{Politecnico di Milano, Milano, Italy \\ \email{daniele.calandriello@mail.polimi.it}
\and Tokyo Institute of Technology, Tokyo, Japan \\ \email{\{gang@sg., sugi@\}cs.titech.ac.jp}}

\maketitle

\begin{abstract}
Semi-supervised clustering aims to introduce prior knowledge in the
decision process of a clustering algorithm.
In this paper, we propose a novel semi-supervised clustering algorithm
based on the information-maximization principle.
The proposed method is an extension of a previous unsupervised 
information-maximization clustering algorithm
based on squared-loss mutual information
to effectively incorporate must-links and cannot-links.
The proposed method is computationally efficient because
the clustering solution can be obtained analytically
via eigendecomposition.
Furthermore, the proposed method allows systematic
optimization of tuning parameters such as the kernel width,
given the degree of belief in the must-links and cannot-links.
The usefulness of the proposed method is demonstrated through experiments.
\end{abstract}

\section{Introduction}\label{Introduction}

The objective of clustering is to classify unlabeled data into disjoint groups
based on their similarity, and clustering has been extensively
studied in statistics and machine learning.
\emph{K-means} \cite{BerkeleySymp:MacQueen:1967}
is a classic algorithm that clusters data so that
the sum of within-cluster scatters is minimized.
However, its usefulness is rather limited in practice
because k-means only produces linearly separated clusters.
\emph{Kernel k-means} \cite{IEEE-TNN:Girolami:2002}
overcomes this limitation by
performing k-means in a feature space induced by a reproducing kernel function
\cite{book:Schoelkopf+Smola:2002}.
\emph{Spectral clustering} \cite{IEEE-PAMI:Shi+Malik:2000,nips02-AA35}
first unfolds non-linear data manifolds 
based on sample-sample similarity by a spectral embedding method,
and then performs k-means in the embedded space.

These non-linear clustering techniques is capable of handling
highly complex real-world data.
However, they lack objective model selection strategies,
i.e., tuning parameters included in kernel functions or similarity measures
need to be manually determined in an unsupervised manner.
\emph{Information-maximization clustering}
can address the issue of model selection
\cite{NIPS2005_569,NIPS2010_0457,ICML:Sugiyama+etal:2011},
which learns a probabilistic classifier so that
some information measure between feature vectors and cluster assignments is maximized
in an unsupervised manner.
In the information-maximization approach,
tuning parameters included in kernel functions or similarity measures
can be systematically determined based on the information-maximization principle.
Among the information-maximization clustering methods,
the algorithm based on \emph{squared-loss mutual information} (SMI)
was demonstrated to be promising \cite{ICML:Sugiyama+etal:2011},
because it gives the clustering solution analytically via eigendecomposition.

In practical situations, 
additional side information regarding clustering solutions is often provided,
typically in the form of \emph{must-links}  and \emph{cannot-links}:
A set of sample pairs which should belong to the same cluster
and a set of sample pairs which should belong to different clusters, respectively.
Such semi-supervised clustering (which is also known as
clustering with side information) has been shown to be useful in practice
\cite{first_links,Goldberg07dissimilarityin,lane_finding}.
\emph{Spectral learning} \cite{SL}
is a semi-supervised extension of spectral clustering
that enhances the similarity with side information
so that sample pairs tied with must-links have higher similarity
and sample pairs tied with cannot-links have lower similarity.
On the other hand, \emph{constrained spectral clustering} \cite{CSP}
incorporates the must-links and cannot-links
as constraints in the optimization problem.

However, in the same way as unsupervised clustering,
the above semi-supervised clustering methods
suffer from lack of objective model selection strategies
and thus tuning parameters included in similarity measures
need to be determined manually.
In this paper, we extend the unsupervised SMI-based clustering method
to the semi-supervised clustering scenario.
The proposed method, called \emph{semi-supervised SMI-based clustering} (3SMIC),
gives the clustering solution analytically via eigendecomposition
with a systematic model selection strategy.
Through experiments on real-world datasets,
we demonstrate the usefulness of the proposed 3SMIC algorithm.

\section{Information-Maximization Clustering with Squared-Loss Mutual Information}\label{IMCSMI}
In this section, we formulate the problem of information-maximization clustering
and review an existing unsupervised clustering method based on squared-loss mutual information.

\subsection{Information-Maximization Clustering}\label{IMC}
The goal of unsupervised clustering is to assign class labels to data instances
so that similar instances share the same label and dissimilar instances have different labels.
Let $\{ \boldx_i|\boldx_i \in \mathbbR^\inputdim\}_{i=1}^\nsample$
be feature vectors of data instances,
which are drawn independently from a probability distribution with density \(\density(\boldx)\).
Let $\{ y_i|y_i \in \{1,\ldots,c\} \}_{i=1}^\nsample$ be class labels
that we want to obtain,
where \(c\) denotes the number of classes and we assume \(c\) to be known through the paper.

The information-maximization approach tries to learn
the class-posterior probability \(\density(y|\boldx)\) in an unsupervised manner
so that some ``information'' measure between feature $\boldx$ and label $y$
is maximized.
\emph{Mutual information} (MI) \cite{Bell:Shannon:1948} is a typical information measure
for this purpose \cite{NIPS2005_569,NIPS2010_0457}:
\begin{align}
  \mathrm{MI}:=\int\sum_{y=1}^{\nclass} \density(\boldx,y)
  \log\frac{\density(\boldx,y)}{\density(\boldx)\density(y)}
  \mathrm{d}\boldx.
\label{equation:MI}
\end{align}
An advantage of the information-maximization formulation is that
tuning parameters included in clustering algorithms such as
the Gaussian width and the regularization parameter can be objectively
optimized based on the same information-maximization principle.
However, MI is known to be sensitive to outliers \cite{Biometrika:Basu+etal:1998},
due to the log function that is strongly non-linear.
Furthermore, unsupervised learning of class-posterior probability \(\density(y|\boldx)\)
under MI is highly non-convex and finding a good local optimum is 
not straightforward in practice \cite{NIPS2010_0457}.

To cope with this problem,
an alternative information measure called \emph{squared-loss MI} (SMI) has been introduced
\cite{BMCBio:Suzuki+etal:2009}:
\begin{align}
  \mathrm{SMI}&:=\frac{1}{2}\int\sum_{y=1}^{\nclass} \density(\boldx)\density(y)
  \left(\frac{\density(\boldx,y)}{\density(\boldx)\density(y)}-1\right)^2
  \mathrm{d}\boldx.
  \label{equation:SMI}
\end{align}
Ordinary MI is the \emph{Kullback-Leibler (KL) divergence}
\cite{Annals-Math-Stat:Kullback+Leibler:1951}
from $\density(\boldx,y)$ to $\density(\boldx)\density(y)$,
while SMI is the \emph{Pearson (PE) divergence} \cite{PhMag:Pearson:1900}.
Both KL and PE divergences belong to 
the class of the \emph{Ali-Silvey-Csisz\'ar divergences}
\cite{JRSS-B:Ali+Silvey:1966,SSM-Hungary:Csiszar:1967},
which is also known as the \emph{\(f\)-divergences}.
Thus, MI and SMI share many common properties, for example they are non-negative and
equal to zero if and only if feature vector $\boldx$ and label $y$ are statistically independent.
Information-maximization clustering based on SMI was shown to be
computationally advantageous \cite{ICML:Sugiyama+etal:2011}.
Below, we review the SMI-based clustering (SMIC) algorithm.

\subsection{SMI-Based Clustering}\label{SMIC}
In unsupervised clustering, it is not straightforward
to approximate SMI \eqref{equation:SMI}
because labeled samples are not available.
To cope with this problem,
let us expand the squared term in Eq.\eqref{equation:SMI}.
Then SMI can be expressed as
\begin{align}
  \mathrm{SMI}
  &=
  \frac{1}{2}\int\sum_{y=1}^{\nclass} \density(\boldx)\density(y)
  \left(\frac{\density(\boldx,y)}{\density(\boldx)\density(y)}\right)^2
  \mathrm{d}\boldx\nonumber\\
  &\phantom{=}
  -\int\sum_{y=1}^{\nclass} \density(\boldx)\density(y)
  \frac{\density(\boldx,y)}{\density(\boldx)\density(y)}
  \mathrm{d}\boldx+\frac{1}{2}\nonumber\\
  &=
  \frac{1}{2}\int\sum_{y=1}^{\nclass} \density(y|\boldx)\density(\boldx)
  \frac{\density(y|\boldx)}{\density(y)}
  \mathrm{d}\boldx -\frac{1}{2}.
  \label{SMI4}
\end{align}
Suppose that the class-prior probability \(\density(y)\) is uniform, i.e.,
\begin{align*}
  p(y)=\frac{1}{\nclass}\mbox{ for }y=1,\ldots,\nclass.
\end{align*}
Then we can express Eq.\eqref{SMI4} as
\begin{align}
  \frac{\nclass}{2}\int\sum_{y=1}^{\nclass}\density(y|\boldx)\density(\boldx)
  \density(y|\boldx)
  \mathrm{d}\boldx -\frac{1}{2}.
  \label{SMI2}
\end{align}

Let us approximate the class-posterior probability $\density(y|\boldx)$ by
the following kernel model:
\begin{align}
  \densitymodel(y|\boldx;\boldalpha)&:=\sum_{i=1}^{\nsample}\alpha_{y,i} K_{}(\boldx,\boldx_i),
  \label{kernel-model}
\end{align}
where $\boldalpha=(\alpha_{1,1},\ldots,\alpha_{\nclass,\nsample})^\top\in\mathbbR^{\nclass\nsample}$
is the parameter vector,
$^\top$ denotes the transpose,
and $K_{}(\boldx,\boldx')$ denotes a kernel function.
Let \(\boldK\) be the kernel matrix whose $(i,j)$ element is given by \(K_{}(\boldx_i,\boldx_j)\)
and let \(\boldalpha_y=(\alpha_{y,1},\ldots,\alpha_{y,\nsample})^\top\in\mathbbR^{\nsample}\).
Approximating the expectation over \(\density(\boldx)\) in Eq.\eqref{SMI2}
with the empirical average of samples \(\{\boldx_i\}_{i=1}^{\nsample}\)
and replacing the class-posterior probability $\density(y|\boldx)$ with
the kernel model $\densitymodel(y|\boldx;\boldalpha)$,
we have the following SMI approximator:
\begin{align}
  \widehat{\mathrm{SMI}}
  &:=
  \frac{\nclass}{2\nsample}\sum_{y=1}^{\nclass}\boldalpha_y^\top\boldK^2\boldalpha_y
  -\frac{1}{2}.
  \label{SMIhat}
\end{align}

Under orthonormality of \(\{\boldalpha_y\}_{y=1}^{\nclass}\),
a global maximizer is given by 
the normalized eigenvectors $\boldphi_1,\ldots,\boldphi_\nclass$ associated with the eigenvalues
$\lambda_1\ge\cdots\ge\lambda_\nsample\ge0$ of $\boldK$.
Because the sign of eigenvector $\boldphi_y$ is arbitrary, we set the sign as
\begin{align*}
  \widetilde{\boldphi}_y=\boldphi_y\times\mathrm{sign}(\boldphi_y^\top\boldone_{\nsample}),
\end{align*}
where $\mathrm{sign}(\cdot)$ denotes the sign of a scalar
and $\boldone_{\nsample}$ denotes the $\nsample$-dimensional vector with all ones.
On the other hand, since
\begin{align*}
  \density(y)=\int\density(y|\boldx)\density(\boldx)\mathrm{d}\boldx
  \approx\frac{1}{\nsample}\sum_{i=1}^\nsample \densitymodel(y|\boldx_i;\boldalpha)
  =\boldalpha_y^\top\boldK\boldone_{\nsample},
\end{align*}
and the class-prior probability was set to be uniform,
we have the following normalization condition:
\begin{align*}
  \boldalpha_y^\top\boldK\boldone_{\nsample}=\frac{1}{\nclass}.
\end{align*}
Furthermore, negative outputs are rounded up to zero
to ensure that outputs are non-negative.

Taking these post-processing issues into account,
cluster assignment $y_i$ for $\boldx_i$ is determined as
the maximizer of the approximation of $p(y|\boldx_i)$:
\begin{align*}
  y_i&=\argmax_y
  \frac{[\max(\boldzero_{\nsample},\boldK\widetilde{\boldphi}_y)]_i}
  {\nclass~\max(\boldzero_{\nsample},\boldK\widetilde{\boldphi}_y)^\top\boldone_{\nsample}}
  =\argmax_y
  \frac{[\max(\boldzero_{\nsample},\widetilde{\boldphi}_y)]_i}
  {\max(\boldzero_{\nsample},\widetilde{\boldphi}_y)^\top\boldone_{\nsample}},
\end{align*}
where
$\boldzero_{\nsample}$ denotes the $\nsample$-dimensional vector with all zeros,
the max operation for vectors is applied in the element-wise manner,
and $[\cdot]_i$ denotes the $i$-th element of a vector.
Note that $\boldK\widetilde{\boldphi}_y=\lambda_y\widetilde{\boldphi}_y$
is used in the above derivation.

For out-of-sample prediction, cluster assignment $y'$
for new sample $\boldx'$ may be obtained as
\begin{align}
  y'&:=
  \argmax_y
  \frac{\max\left(0,\sum_{i=1}^\nsample K(\boldx',\boldx_i)[\widetilde{\boldphi}_y]_i\right)}
  {\lambda_y\max(\boldzero_{\nsample},\widetilde{\boldphi}_y)^\top\boldone_{\nsample}}.\label{SMIC:OutOfSample}
\end{align}
This clustering algorithm is called the \emph{SMI-based clustering} (SMIC).

SMIC may include a tuning parameter, say $\theta$, in the kernel function,
and the clustering results of SMIC depend on the choice of $\theta$.
A notable advantage of information-maximization clustering is that
such a tuning parameter can be systematically optimized
by the same information-maximization principle.
More specifically, 
cluster assignments $\{y^{\theta}_i\}_{i=1}^\nsample$ are first obtained for each possible $\theta$.
Then the quality of clustering is measured by the SMI value
estimated from paired samples $\{(\boldx_i,y^{\theta}_i)\}_{i=1}^\nsample$.
For this purpose, the method of \emph{least-squares mutual information} (LSMI)
\cite{BMCBio:Suzuki+etal:2009} is useful
because LSMI was theoretically proved to be the optimal non-parametric SMI approximator
\cite{NC:Suzuki+Sugiyama:2013}; see Appendix~\ref{app:LSMI} for the details of LSMI.
Thus, we compute LSMI as a function of $\theta$
and the tuning parameter value that maximizes LSMI is selected
as the most suitable one:
\begin{align*}
  \max_\theta \mbox{LSMI}(\theta)
\end{align*}

\section{Semi-Supervised SMIC}\label{3SMIC}
In this section, we extend SMIC to a semi-supervised clustering scenario
where
a set of \emph{must-links}  and a set of \emph{cannot-links} are provided.
A must-link $(i,j)$ means that $\boldx_i$ and $\boldx_j$ are encouraged to
belong to the same cluster,
while a cannot-link $(i,j)$ means that $\boldx_i$ and $\boldx_j$ are encouraged to
belong to different clusters.
Let $\boldM$ be the must-link matrix with $M_{i,j}=1$ if a must-link between
$\boldx_i$ and $\boldx_j$ is given and $M_{i,j}=0$ otherwise.
In the same way, we define the cannot-link matrix $\boldC$.
We assume that $M_{i,i}=1$ for all $i=1,\ldots,\nsample$,
and $C_{i,i}=0$ for all $i=1,\ldots,\nsample$.
Below, we explain how must-link constraints and cannot-link constraints
are incorporated into the SMIC formulation.

\subsection{Incorporating Must-Links in SMIC}
When there exists a must-link between $\boldx_i$ and $\boldx_j$,
we want them to share the same class label.
Let 
\begin{align*}
  \boldp^*_i=(\density(y=1|\boldx_i),\ldots,\density(y=\nclass|\boldx_i))^\top
\end{align*}
be the soft-response vector for $\boldx_i$.
Then the inner product $\langle\boldp^*_i,\boldp^*_j\rangle$ is maximized
if and only if
$\boldx_i$ and $\boldx_j$ belong to the same cluster with perfect confidence,
i.e., $\boldp^*_i$ and $\boldp^*_j$ are the same vector
that commonly has $1$ in one element and $0$ otherwise.
Thus, the must-link information may be utilized
by increasing $\langle\boldp^*_i,\boldp^*_j\rangle$ if $M_{i,j}=1$.
We implement this idea as
\begin{align*}
  &\widehat{\mathrm{SMI}}
  +\gamma\frac{\nclass}{\nsample}\sum_{i,j=1}^\nsample M_{i,j} 
  \sum_{y=1}^{\nclass}
  \densitymodel(y|\boldx_i;\boldalpha)
  \densitymodel(y|\boldx_j;\boldalpha),
\end{align*}
where $\gamma\ge0$ determines how strongly
we encourage the must-links to be satisfied.

Let us further utilize the following fact:
If $\boldx_i$ and $\boldx_j$ belong to the same class
and $\boldx_j$ and $\boldx_k$ belong to the same class,
$\boldx_i$ and $\boldx_k$ also belong to the same class
(i.e., a friend's friend is a friend).
Letting $M'_{i,j}=\sum_{k=1}^\nsample M_{i,k}M_{k,j}$,
we can incorporate this in SMIC as
\begin{align*}
  &\widehat{\mathrm{SMI}}
  +\gamma\frac{\nclass}{\nsample}\sum_{i,j=1}^\nsample M_{i,j} 
  \sum_{y=1}^{\nclass}\densitymodel(y|\boldx_i;\boldalpha)
  \densitymodel(y|\boldx_j;\boldalpha)\\
  &\phantom{\widehat{\mathrm{SMI}}}
  +\gamma'\frac{\nclass}{2\nsample}\sum_{i,j=1}^\nsample M'_{i,j} 
  \sum_{y=1}^{\nclass}\densitymodel(y|\boldx_i;\boldalpha)
  \densitymodel(y|\boldx_j;\boldalpha)\\
  &=  \frac{\nclass}{2\nsample}\sum_{y=1}^{\nclass}\boldalpha_y^\top\boldK^2\boldalpha_y
  -\frac{1}{2}+\gamma\frac{\nclass}{\nsample}\sum_{y=1}^{\nclass}
  \boldalpha_y^\top \boldK \boldM \boldK \boldalpha_y
  +\gamma'\frac{\nclass}{2\nsample}\sum_{y=1}^{\nclass}
  \boldalpha_y^\top \boldK \boldM^2 \boldK \boldalpha_y\\
  &=  \frac{\nclass}{2\nsample}\sum_{y=1}^{\nclass}\boldalpha_y^\top
  \boldK(\boldI+2\gamma\boldM+\gamma'\boldM^2)\boldK\boldalpha_y-\frac{1}{2}.
\end{align*}
If we set $\gamma'=\gamma^2$, we have a simpler form:
\begin{align*}
  \frac{\nclass}{2\nsample}\sum_{y=1}^{\nclass}\boldalpha_y^\top
  \boldK(\boldI+\gamma\boldM)^2\boldK\boldalpha_y-\frac{1}{2},
\end{align*}
which will be used later.

\subsection{Incorporating Cannot-Links in SMIC}\label{3SMIC:CL}
We may incorporate cannot-links in SMIC 
in the opposite way to must-links, 
by decreasing
the inner product $\langle\boldp^*_i,\boldp^*_j\rangle$ to zero.
This may be implemented as
\begin{align*}
  \widehat{\mathrm{SMI}}-\eta\frac{\nclass}{\nsample}\sum_{i,j=1}^\nsample C_{i,j} 
  \sum_{y=1}^{\nclass}
  \densitymodel(y|\boldx_i;\boldalpha)
  \densitymodel(y|\boldx_j;\boldalpha),
\end{align*}
where $\eta\ge0$ determines how strongly we encourage the cannot-links to be satisfied.

In binary clustering problems where $\nclass=2$,
if $\boldx_i$ and $\boldx_j$ belong to different classes
and $\boldx_j$ and $\boldx_k$ belong to different classes,
$\boldx_i$ and $\boldx_k$ actually belong to the same class
(i.e., an enemy's enemy is a friend).
Let $C'_{i,j}=\sum_{k=1}^\nsample C_{i,k}C_{k,j}$,
and we will take this also into account as must-links in the following way:
\begin{align*}
  &\widehat{\mathrm{SMI}}-\eta\frac{\nclass}{\nsample}\sum_{i,j=1}^\nsample C_{i,j} 
  \sum_{y=1}^{\nclass}
  \densitymodel(y|\boldx_i;\boldalpha)
  \densitymodel(y|\boldx_j;\boldalpha)\\
  &\phantom{\widehat{\mathrm{SMI}}}
  +\eta'\frac{\nclass}{2\nsample}
  \sum_{i,j=1}^\nsample C'_{i,j} 
  \sum_{y=1}^{\nclass}
  \densitymodel(y|\boldx_i;\boldalpha)
  \densitymodel(y|\boldx_j;\boldalpha)\\
  &=\frac{\nclass}{2\nsample}\sum_{y=1}^{\nclass}\boldalpha_y^\top
  \boldK(\boldI-2\eta\boldC+\eta'\boldC^2)\boldK\boldalpha_y-\frac{1}{2}.
\end{align*}
If we set $\eta'=\eta^2$, we have
\begin{align*}
  &\frac{\nclass}{2\nsample}\sum_{y=1}^{\nclass}\boldalpha_y^\top
  \boldK(\boldI-\eta\boldC)^2\boldK\boldalpha_y-\frac{1}{2},
\end{align*}
which will be used later.

\subsection{Kernel Matrix Modification}
Another approach to incorporating must-links and cannot-links
is to modify the kernel matrix $\boldK$.
More specifically,
$K_{i,j}$ is increased
if there exists a must-link between $\boldx_i$ and $\boldx_j$,
and
$K_{i,j}$ is decreased
if there exists a cannot-link between $\boldx_i$ and $\boldx_j$.
In this paper, we assume $K_{i,j}\in[0,1]$,
and set $K_{i,j}=1$ if there exists a must-link between $\boldx_i$ and $\boldx_j$
and $K_{i,j}=0$ if there exists a cannot-link between $\boldx_i$ and $\boldx_j$.
Let us denote the modified kernel matrix by $\boldK'$:
\begin{align*}
  \boldK'\longleftarrow\boldK.
\end{align*}

This modification idea has been employed in spectral clustering \cite{SL} 
and demonstrated to be promising.

\subsection{Semi-Supervised SMIC}\label{3SMIC:together}
Finally, we combine the above three ideas as
\begin{align*}
  \widetilde{\mathrm{SMI}}:=
  \frac{\nclass}{2\nsample}\sum_{y=1}^{\nclass}\boldalpha_y^\top
  \boldU
  \boldalpha_y-\frac{1}{2},
\end{align*}
where
\begin{align}
  \boldU:=\boldK'
  (2\boldI+2\gamma\boldM+\gamma^2\boldM^2-2\eta\boldC+\eta^2\boldC^2)
  \boldK'
  \label{boldU}
\end{align}
When $\nclass>2$, we fix $\eta$ at zero.

This is the learning criterion of \emph{semi-supervised SMIC} (3SMIC),
whose global maximizer can be analytically obtained
under orthonormality of \(\{\boldalpha_y\}_{y=1}^{\nclass}\)
by the leading eigenvectors of $\boldU$.
Then the same post-processing as the original SMIC is applied
and cluster assignments are obtained.
Out-of-sample prediction is also possible in the same way as the original SMIC.

\subsection{Tuning Parameter Optimization in 3SMIC}
In the original SMIC, an SMI approximator called LSMI is used
for tuning parameter optimization (see Appendix~\ref{app:LSMI}).
However, this is not suitable in semi-supervised scenarios
because the 3SMIC solution is biased to satisfy must-links and cannot-links.
Here, we propose using
\begin{align*}
\max_\theta \mbox{LSMI}(\theta) + \mbox{Penalty}(\theta),
\end{align*}
where $\theta$ indicates tuning parameters in 3SMIC;
in the experiments, $\gamma$, $\eta$, and
the parameter $\kernelparameter$ included in the kernel function $K(\boldx,\boldx')$
is optimized.
``$\mbox{Penalty}$'' is the penalty for violating must-links and cannot-links,
which is the only tuning factor in the proposed algorithm.

\section{Experiments}\label{EXP:TEMP}
In this section, we experimentally evaluate
the performance of the proposed 3SMIC method
in comparison with popular semi-supervised clustering methods:
\emph{Spectral Learning} (SL) \cite{SL} and
\emph{Constrained Spectral Clustering} (CSC) \cite{CSP}.
Both methods first perform semi-supervised spectral embedding 
and then k-means to obtain clustering results.
However, we observed that the post k-means step is often unreliable,
so we use simple thresholding \cite{IEEE-PAMI:Shi+Malik:2000}
in the case of binary clustering for CSC.

In all experiments, we will use
a sparse version of the \emph{local-scaling kernel} \cite{NIPS17:Zelnik-Manor+Perona:2005}
as the similarity measure:
\begin{align*}
  K(\boldx_i,\boldx_j)
  =
  \begin{cases}
    \displaystyle
    \exp\left(-\frac{\|\boldx_i-\boldx_j\|^2}{2\sigma_i\sigma_j}\right)
    & \mbox{if $\boldx_i\in \calN_\kernelparameter(\boldx_j)$
      or $\boldx_j\in \calN_\kernelparameter(\boldx_i)$},\\
    0 & \mbox{otherwise},
  \end{cases}
\end{align*}
where $\calN_\kernelparameter(\boldx)$ denotes the set of
$\kernelparameter$ nearest neighbors for $\boldx$
($\kernelparameter$ is the kernel parameter),
$\sigma_i$ is a local scaling factor defined as
$\sigma_i=\|\boldx_i-\boldx_i^{(\kernelparameter)}\|$,
and $\boldx_i^{(\kernelparameter)}$ is
the $\kernelparameter$-th nearest neighbor of $\boldx_i$.
For SL and CSC, we test $\kernelparameter=1,4,7,10$
(note that there is no systematic way to choose the value of $\kernelparameter$),
except for the \textbf{spam} dataset with \(\kernelparameter=1\)
that caused numerical problems in the eigensolver when testing SL.
On the other hand, in 3SMIC,
we choose the value of $\kernelparameter$ from $\{1,\ldots,10\}$ 
based on the following criterion:
\begin{align}
   &\frac{\mbox{LSMI}(\theta)}{\max_\theta\mbox{LSMI}(\theta)}
   - \frac{n_v}{\max_\theta(n_v)},
\label{EXP:PriorPenalty}
\end{align}
where
$n_v$ is the number of violated links. 
Here, both the LSMI value and the penalty are normalized so that they fall into the range $[0,1]$.
The \(\gamma\) and \(\eta\) parameters in 3SMIC 
are also chosen based on Eq.\eqref{EXP:PriorPenalty}.

We use the following real-world datasets:
\begin{description}
  \item[parkinson] ($d=22$, $n=195$, and $c=2$): The UCI dataset
    consisting of voice registration 
  from patients suffering Parkinson's disease and sane individuals.
  From the voice, 22 feature are extracted.

  \item[spam] ($d=57$, $n=4601$, and $c=2$): The UCI dataset consisting of e-mails,
    categorized in spam and non-spam.
    48 word-frequency features and 9 other frequency features
    such as specific characters and capitalization are extracted.

  \item[sonar] ($d=60$, $n=208$, and  $c=2$): The UCI dataset consisting of sonar responses
  from a metal object or a rock. The features represent energy in each frequency band.

  \item[digits500] ($d=256$, $n=500$, and $c=10$): The 
  USPS digits dataset consisting of images of written numbers from 0 to 9, 
  256 $(16\times16)$ pixels in gray-scale. We randomly sampled 50 numbers 
  for each digit, and normalized each pixel intensity in the image between $-1$ and $1$.

  \item[digits5k] ($d=256$, $n=5000$, and $c=10$): The same USPS digits dataset but with
  500 images for each class.

  \item[faces100] ($d=4096$, $n=100$, and $c=10$): 
    The Olivetti Face dataset consisting of images of human faces
  in gray-scale, 4096 $(64\times 64)$ pixels. 
  We randomly selected 10 persons, and used 10 images for each person.

\end{description}

Must-links and cannot-links are generated from the true labels,
by randomly sampling a couple of points
and adding the corresponding 1 to
the \(\boldM\) or \(\boldC\) matrices depending on the labels of the chosen pair of points.
CSC is excluded from \textbf{digits5k} and \textbf{spam}
because it needs to solve the complete eigenvalue problem
and its computational cost was too high on these large datasets.

\begin{figure}[p]
\centering
\includegraphics[width=12.2cm]{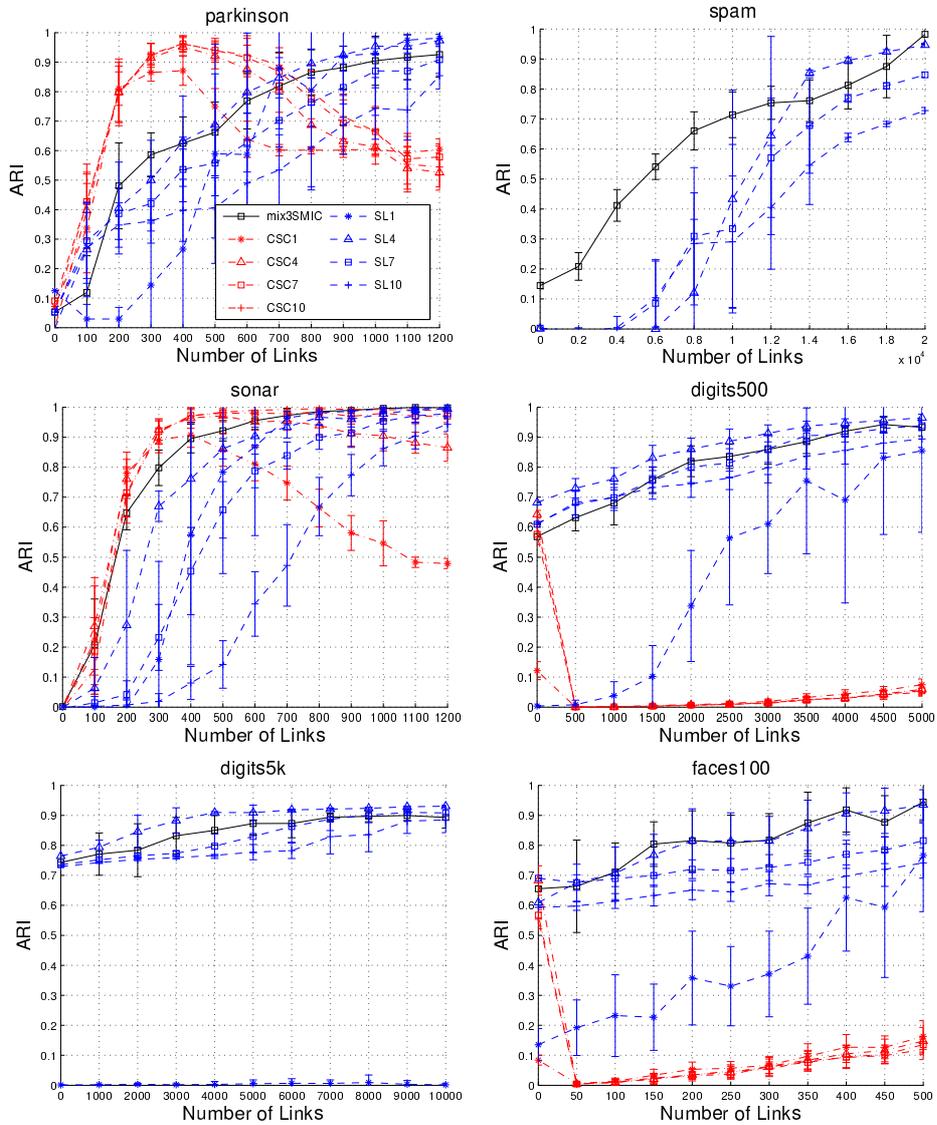}
\caption{Experimental results.}
\label{EXP:real_ds}
\end{figure}

We evaluate the clustering performance by the \emph{Adjusted Rand Index} (ARI)
\cite{JoC:Hubert+Arabie:1985} between learned and true labels.
Larger ARI values mean better clustering performance,
and the zero ARI value means that the clustering result is equivalent to random.
We investigate the ARI score as functions of the number of links used.
Averages and standard deviations of ARI over $20$ runs with different
random seeds are plotted in Figure~\ref{EXP:real_ds}.

We can separate the datasets into two groups.
For \textbf{digits500}, \textbf{digits5k}, and \textbf{faces100}, 
the baseline performances without links are reasonable;
the introduction of links significantly increase the performance,
bringing it around $0.9$--$0.95$ from $0.5$--$0.8$.

For \textbf{parkinson}, \textbf{spam}, and \textbf{sonar}
where the baseline performances without links are poor,
introduction of links quickly allow the clustering algorithms
to find better solutions.
In particular, only 3\% of links (relative to all possible pairs)
was sufficient for \textbf{parkinson} to achieve reasonable performance
and surprisingly only 0.1\% for \textbf{spam}.

As shown in Figure~\ref{EXP:real_ds},
the performance of SL depends heavily on the choice of $\kernelparameter$,
but there is no systematic way to choose $\kernelparameter$ for SL.
It is important to notice that 3SMIC with $\kernelparameter$ 
chosen systematically based on Eq.\eqref{EXP:PriorPenalty}
performs as good as SL with $\kernelparameter$ tuned optimally with hindsight.
On the other hand,
CSC performs rather stably for different values of $\kernelparameter$,
and it works particularly well for binary problems with a small number of links.
However, it performs very poorly for multi-class problems;
we observed that the post k-means step is highly unreliable
and poor local optimal solutions are often produced.
For the binary problems, simply performing thresholding \cite{IEEE-PAMI:Shi+Malik:2000}
instead of using k-means was found to be useful.
However, there seems no simple alternatives in multi-class cases.
The performance of CSC drops in \textbf{parkinson} and \textbf{sonar} 
when the number of links is increased,
although such phenomena were not observed in SL and 3SMIC.

Overall, the proposed 3SMIC method was shown to be a promising 
semi-supervised clustering method.

\section{Conclusions}\label{sec:conclusions}
In this paper, we proposed a novel information-maximization clustering method
that can utilize side information provided as must-links and cannot-links.
The proposed method, named \emph{semi-supervised SMI-based clustering} (3SMIC),
allows us to compute the clustering solution
analytically. This is a strong advantage over conventional approaches
such as \emph{constrained spectral clustering} (CSC)
that requires a post k-means step,
because this post k-means step can be unreliable and cause
significant performance degradation in practice.
Furthermore, 
3SMIC allows us to systematically determine tuning parameters
such as the kernel width based on the information-maximization
principle, given our reliance on the provided side information.
Through experiments, we demonstrated that automatically-tuned
3SMIC perform as good as optimally-tuned \emph{spectral learning} (SL) with hindsight.

The focus of our method in this paper was to inherit
the analytical treatment of the original unsupervised SMIC
in semi-supervised learning scenarios.
Although this analytical treatment was demonstrated to be highly useful
in experiments,
our future work will explore more efficient use of must-links and cannot-links.

In the previous work \cite{JMLR:Laub+Mueller:2004},
negative eigenvalues were found to contain useful information.
Because must-link and cannot-link matrices can possess negative eigenvalues,
it is interesting to investigate the role and effect of
negative eigenvalues in the context of information-maximization clustering.

\section*{Acknowledgements}
This work was carried out when DC was visiting at Tokyo Institute of Technology
by the YSEP program.
GN was supported by the MEXT scholarship,
and MS was supported by MEXT KAKENHI 25700022 and AOARD.

\appendix
\section{Least-Squares Mutual Information}
\label{app:LSMI}
The solution of SMIC depends on the choice of
the kernel parameter 
included in the kernel function $K(\boldx,\boldx')$.
Since SMIC was developed in the framework of SMI maximization,
it would be natural to determine the kernel parameter
so as to maximize SMI.
A direct approach is to use the SMI estimator $\widehat{\mathrm{SMI}}$
given by Eq.\eqref{SMIhat}
also for kernel parameter choice.
However, this direct approach is not favorable
because $\widehat{\mathrm{SMI}}$ is an unsupervised SMI estimator
(i.e., SMI is estimated only from unlabeled samples $\{\boldx_i\}_{i=1}^{\nsample}$).
On the other hand, in the model selection stage, we have already obtained
labeled samples $\{(\boldx_i,y_i)\}_{i=1}^{\nsample}$,
and thus supervised estimation of SMI is possible.
For supervised SMI estimation, a non-parametric SMI estimator
called \emph{least-squares mutual information} (LSMI) \cite{BMCBio:Suzuki+etal:2009}
was proved to achieve the optimal convergence rate to the true SMI.
Here we briefly review LSMI.

The key idea of LSMI is to learn the following \emph{density-ratio function}
\cite{book:Sugiyama+etal:2012},
\begin{align*}
\ratio(\boldx,y) := \frac{\density(\boldx,y)}{\density(\boldx)\density(y)},
\end{align*}
without going through probability density/mass estimation of
$\density(\boldx,y)$, $\density(\boldx)$, and $\density(y)$.
More specifically,
let us employ the following density-ratio model:
\begin{align}
\ratiomodel(\boldx,y;\boldomega) &:=\sum_{\ell:y_\ell=y}\omega_{\ell}
L(\boldx,\boldx_\ell),
\label{ratio-model}
\end{align}
where $\boldomega=(\omega_1,\ldots,\omega_{\nsample})^\top$
and
$L(\boldx,\boldx')$ is a kernel function.
In practice, we use the Gaussian kernel
\begin{align*}
  L(\boldx,\boldx')=
  \exp\left(-\frac{\|\boldx-\boldx'\|^2}{2\kappa^2}\right),
\end{align*}
where the Gaussian width $\kappa$ is the kernel parameter.
To save the computation cost, we limit the number of kernel bases
to $500$ with randomly selected kernel centers.

The parameter $\boldomega$ in the above density-ratio model is learned so that
the following squared error is minimized:
\begin{align}
  \min_\boldomega
  \frac{1}{2}\int\sum_{y=1}^{\nclass}\Big(\ratiomodel(\boldx,y;\boldomega)-\ratio(\boldx,y)\Big)^2
  \density(\boldx)\density(y)\mathrm{d}\boldx.
  \label{LSMI-squared-error}
\end{align}
Let $\boldomega^{(y)}$ be the parameter vector
corresponding to the kernel bases
$\{L(\boldx,\boldx_\ell)\}_{\ell:y_\ell=y}$,
i.e., $\boldomega^{(y)}$ is the sub-vector of $\boldomega=(\omega_1,\ldots,\omega_\nsample)^\top$
consisting of indices $\{\ell\;|\;y_\ell=y\}$.
Let $\nsample_y$ be the number of samples in class $y$,
which is the same as the dimensionality of $\boldomega^{(y)}$.
Then an empirical and regularized version
of the optimization problem \eqref{LSMI-squared-error} is given
for each $y$ as follows:
\begin{align}
  \min_{\boldomega^{(y)}}
  \left[
    \frac{1}{2}\boldomega^{(y)}{}^\top\boldHh^{(y)}\boldomega^{(y)}
  -\boldomega^{(y)}{}^\top\boldhh^{(y)}
  +\frac{\delta}{2}\boldomega^{(y)}{}^\top\boldomega^{(y)}
  \right],
\label{LSMI-objective}
\end{align}
where $\delta$ ($\ge0$) is the regularization parameter.
$\boldHh^{(y)}$ is the $\nsample_y\times\nsample_y$ matrix
and $\boldhh^{(y)}$ is the $\nsample_y$-dimensional vector defined as
\begin{align*}
\Hh_{\ell,\ell'}^{(y)}&:=\frac{\nsample_y}{\nsample^2}
\sum_{i=1}^\nsample L(\boldx_i,\boldx_{\ell}^{(y)})
L(\boldx_i,\boldx_{\ell'}^{(y)}),~~~
\hh_{\ell}^{(y)}:=\frac{1}{\nsample}
\sum_{i:y_i=y}L(\boldx_i,\boldx_{\ell}^{(y)}),
\end{align*}
where $\boldx^{(y)}_\ell$ is the $\ell$-th sample in class $y$
(which corresponds to $\omegah^{(y)}_{\ell}$).

A notable advantage of LSMI is that
the solution $\boldomegah^{(y)}$ can be computed analytically as
\begin{align*}
  \boldomegah^{(y)}=(\boldHh^{(y)}+\delta\boldI)^{-1}\boldhh^{(y)}.
\end{align*}
Then a density-ratio estimator is obtained analytically as follows:
\begin{align*}
\ratioh(\boldx,y)&=\sum_{\ell=1}^{\nsample_y}\omegah^{(y)}_{\ell}
L(\boldx,\boldx^{(y)}_\ell).
\end{align*}

The accuracy of the above least-squares density-ratio estimator depends on
the choice of the kernel parameter $\kappa$ included in $L(\boldx,\boldx')$
and the regularization parameter $\delta$ in Eq.\eqref{LSMI-objective}.
These tuning parameter values
can be systematically optimized based on cross-validation as follows:
First, the samples $\calZ=\{(\boldx_i,y_i)\}_{i=1}^{\nsample}$
are divided into $M$ disjoint subsets $\{\calZ_m\}_{m=1}^M$
of approximately the same size (we use $M=5$ in the experiments).
Then a density-ratio estimator $\ratioh_m(\boldx,y)$
is obtained using $\calZ\backslash\calZ_m$
(i.e., all samples without $\calZ_m$),
and its out-of-sample error (which corresponds to Eq.\eqref{LSMI-squared-error}
without irrelevant constant)
for the hold-out samples $\calZ_m$ is computed as
\begin{align*}
  \mathrm{CV}_m:=\frac{1}{2|\calZ_m|^2}\sum_{\boldx,y\in\calZ_m}\ratioh_m(\boldx,y)^2
  -\frac{1}{|\calZ_m|}\sum_{(\boldx,y)\in\calZ_m}\ratioh_m(\boldx,y),
\end{align*}
where $\sum_{\boldx,y\in\calZ_m}$ denotes the summation over all combinations of
$\boldx$ and $y$ in $\calZ_m$ (and thus $|\calZ_m|^2$ terms),
while $\sum_{(\boldx,y)\in\calZ_m}$ denotes the summation over all pairs
$(\boldx,y)$ in $\calZ_m$ (and thus $|\calZ_m|$ terms).
This procedure is repeated for $m=1,\ldots,M$,
and the average of the above hold-out error
over all $m$ is computed as
\begin{align*}
  \mathrm{CV}:=\frac{1}{M}\sum_{m=1}^M  \mathrm{CV}_m.
\end{align*}
Then the kernel parameter $\kappa$
and the regularization parameter $\delta$
that minimize the average hold-out error $\mathrm{CV}$
are chosen as the most suitable ones.

Finally, given that SMI \eqref{equation:SMI} can be expressed as
\begin{align*}
  \mathrm{SMI}&=
  -\frac{1}{2}\int\sum_{y=1}^{\nclass}\ratio(\boldx,y)^2\density(\boldx)\density(y)\mathrm{d}\boldx
+ \int\sum_{y=1}^{\nclass}\ratio(\boldx,y)\density(\boldx,y)\mathrm{d}\boldx
-\frac{1}{2},
\end{align*}
an SMI estimator based on the above density-ratio estimator,
called \emph{least-squares mutual information} (LSMI), is given as follows:
\begin{align*}
  \mathrm{LSMI}:=
  -\frac{1}{2\nsample^2}\sum_{i,j=1}^{\nsample}\ratioh(\boldx_i,y_j)^2
    +\frac{1}{\nsample}\sum_{i=1}^{\nsample}\ratioh(\boldx_i,y_i)-\frac{1}{2},
\end{align*}
where $\ratioh(\boldx,y)$ is a density-ratio estimator obtained above.

\bibliographystyle{plainnat}
\bibliography{./reducedbib}
\end{document}